\documentclass[runningheads]{llncs}
\usepackage{graphicx}
\usepackage{amsmath,amssymb} 
\usepackage{color}
\usepackage{amsmath}
\usepackage{amssymb}
\usepackage{floatrow}
\usepackage{subfig}
\usepackage{wrapfig}
\usepackage{cuted}
\usepackage{blindtext}
\usepackage{bbm}
\usepackage{diagbox}
\usepackage{multirow}
\usepackage[font=small,labelfont=bf,tableposition=top]{caption}
\DeclareCaptionLabelFormat{andtable}{#1~#2  \&  \tablename~\thetable}

\DeclareMathOperator*{\argmin}{argmin}
\newcommand{\tabincell}[2]{\begin{tabular}{@{}#1@{}}#2\end{tabular}}

\begin{document}
\title{GridFace: Face Rectification via Learning Local Homography Transformations} 

\titlerunning{Face Rectification with Homographies}
%
\author{Erjin Zhou \and Zhimin Cao \and Jian Sun}
%
\authorrunning{E. Zhou and Z. Cao and J. Sun}
%

\institute{Face++, Megvii Inc.\\
\email{\{zej,czm,sunjian\}@megvii.com}}
\maketitle              
\begin{abstract}
In this paper, we propose a method, called GridFace, to reduce facial geometric variations and improve the recognition performance. 
Our method rectifies the face by local homography transformations, which are estimated by a face rectification network. 
To encourage the image generation with canonical views, we apply a regularization based on the natural face distribution. 
We learn the rectification network and recognition network in an end-to-end manner. 
Extensive experiments show our method greatly reduces geometric variations, and gains significant improvements in unconstrained face recognition scenarios.
\end{abstract}
\begin{figure}
\vspace{-1cm}
    \centering\noindent
    \includegraphics[width=0.95\linewidth]{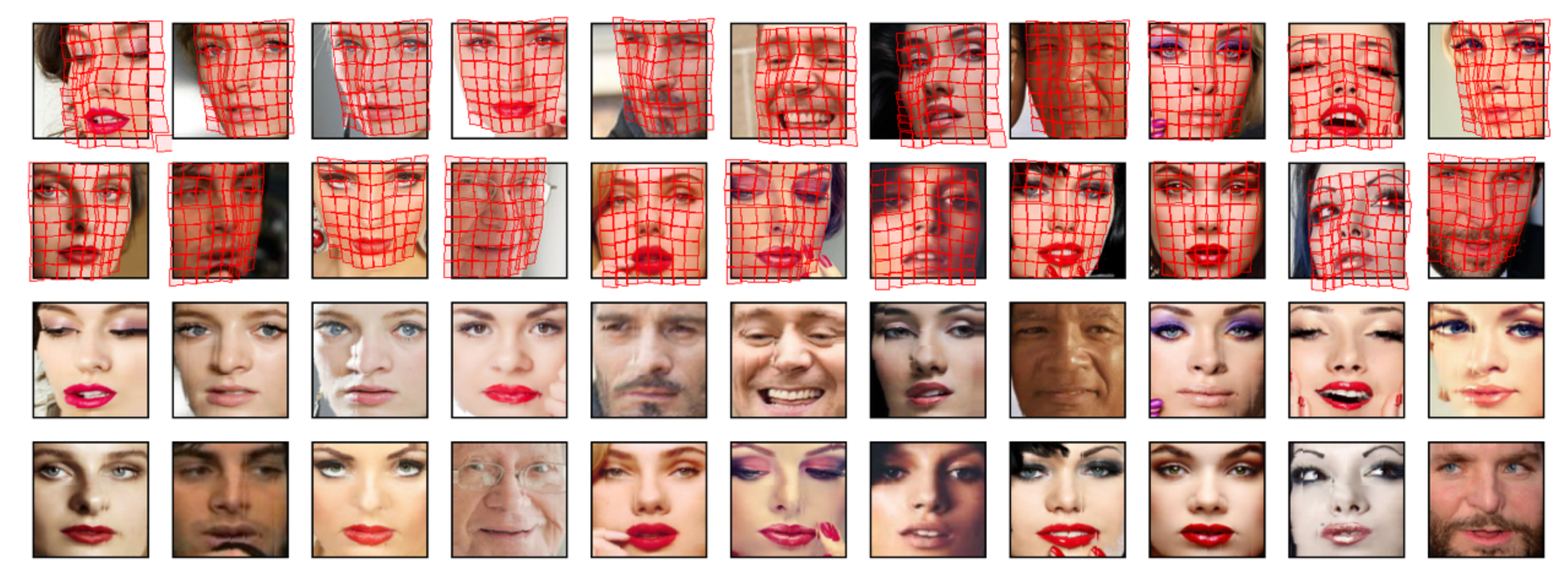}
    \caption{{\bf Face Rectification Examples.}
    The top two rows are faces with large geometric variations, and local homographies estimated by the rectification network. 
    The bottom two rows show the rectified faces by local homographies, which greatly reduce the geometric variations and calibrate the faces into canonical view. 
}
\end{figure}
\vspace{-1.cm}

\section{Introduction}
Despite of the recent academic/commercial progresses made in deep learning~\cite{Taigman_2014_CVPR}, \cite{Sun_2014_CVPR}, \cite{NIPS2014_5416}, \cite{zhu2014multi}, \cite{Yim_2015_CVPR}, \cite{Schroff_2015_CVPR}, \cite{Liu_2017_CVPR}, \cite{wen2016discriminative}, \cite{Masi_2016_CVPR}, \cite{Tran_2017_CVPR}, \cite{Huang_2017_ICCV}, \cite{Yin_2017_ICCV}, \cite{Zhang_2017_ICCV}, \cite{Wu_2017_ICCV}, it is still hard to claim that face recognition has been solved in unconstrained settings. One of the remaining challenges for the in-the-wild recognition is facial geometric variations.
Such variations in pose and misalignment (introduced by face detection bounding box localization) substantially degrade the face representation and recognition performance.

The common adopted way to deal with this issue is using a 2D transformation to calibrate the facial landmarks to pre-defined templates (i.e., 2D mean face landmarks or a 3D mean face model). 
However, such kind of pre-processing is not optimized towards the recognition system and relies heavily on the parameters tuned by hand and accurate facial landmarks. To address this problem, recent works use the Spatial Transformer Network (STN)~\cite{jaderberg2015spatial} to perform an end-to-end optimization with consideration of both face alignment and detection/recognition~\cite{chen2016supervised}, \cite{7948720}.
However, the transformation learned in these works uses a holistic parametric model that can only capture coarse geometric information, such as facial orientation, and may introduce notable distortion in the rectified results. 

\begin{figure}[t]
\begin{center}
   \includegraphics[width=1.0\linewidth]{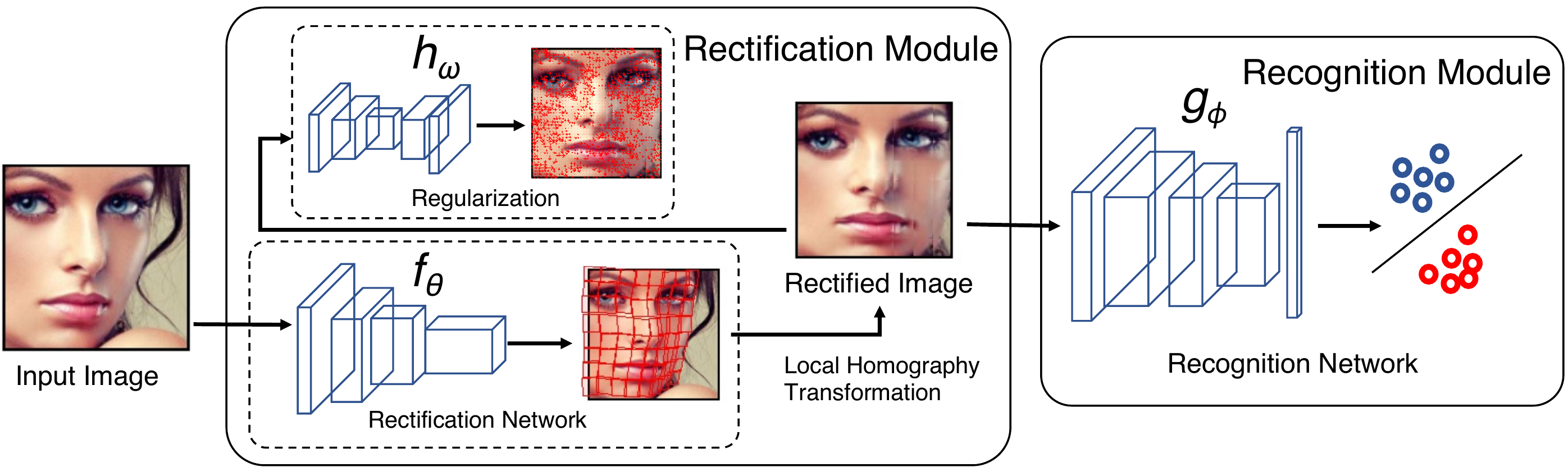}
\end{center}
   \caption{
   \textbf{System Overview.} 
   The system contains two modules: the rectification module and the recognition module.
   The rectification module extracts deep feature by the rectification network and warps the image with a group of local homography transformations (Sec.~\ref{sec:warp}). 
   The rectified output is regularized by an implicit canonical view face prior, which is optimized by a Denoising Autoencoder (Sec.~\ref{sec:prior}).
   The red arrows in the face in the regularization box indicate the approximated gradients estimated by DAE.
   With the rectified faces as input, the recognition network learns discriminative face representation (Sec.~\ref{sec:recog}) via metric learning.
   The whole system is end-to-end optimized with stochastic gradient descent. 
   }
\label{fig:overview}
\end{figure}

In this paper, we propose a novel method called \emph{GridFace} to reduce the facial geometric variations and boost the recognition performance.
As shown in Fig.~\ref{fig:overview}, our system contains two main modules: the rectification module and the recognition module.

In the rectification module, we apply a face rectification network to estimate a group of local homography transformations for rectifying the input facial image (Sec.~\ref{sec:warp}). 
We approximate the underlying 3D canonical face shape by a group of deformable plane cells. 
When a face with geometric variations fed in, local homography transformations are estimated to model the warping of each cell respectively.
In order to encourage the generation with canonical views, we introduce a regularization based on the canonical view face distribution (Sec.~\ref{sec:prior}). 
This natural face distribution is not explicitly modeled. Instead, we use a Denoising Autoencoder (DAE) to estimate the gradients of logarithm of probability density, 
which is inspired by the previous work~\cite{sarela2005denoising,alain2014regularized}. 
The recognition module (Sec.~\ref{sec:recog}) takes the rectified image as input and learns discriminative representation via metric learning.

In Sec.~\ref{sec:exp}, we first evaluate our method with qualitative and quantitative results to demonstrate the effectiveness of face rectification for recognition in-the-wild.
Then we present extensive ablation studies to show the importance of each of the above components.
We finally evaluate our method on four challenging public benchmarks LFW, YTF, IJB-A, and Multi-PIE.
We obtain large improvement in all benchmarks, and achieve superior or comparable results compared with recent face frontalization and recognition works.

Our contributions are summarized as following:
\begin{enumerate}
\item We propose a novel method to improve face recognition performance by reducing facial geometric variations with local homography transformations.
\item We introduce a canonical face prior and a Denoising Autoencoder based approximation method to regularize the face rectification process for better rectification quality.
\item Extensive experiments on constrained and unconstrained environments are conducted to demonstrate the excellent performance of our method.
\end{enumerate}

\section{Related Works}
\noindent \textbf{Deep Face Recognition.}
Early works~\cite{Sun_2014_CVPR}, \cite{Taigman_2014_CVPR} learn face representation by multi-class classification networks.
Features learned from thousands of individuals' faces demonstrate good generalization ability. 
Sun et al.~\cite{NIPS2014_5416} improve the performance by jointly learning identification and verification losses. 
Schroff et al.~\cite{Schroff_2015_CVPR} formulate the representation learning task in a metric learning framework, 
and introduce the triplet loss and hard negative sample mining skill to boost the performance further.
Recent works~\cite{wen2016discriminative}, \cite{Liu_2017_CVPR} propose the center loss and sphere loss to further reduce intra-class variations in the feature space.
Du and Liang~\cite{Du_2015_CVPR} propose age-invariant feature.
Bhattarai et al.~\cite{Bhattarai_2016_CVPR} introduce multitask learning for large scale face retrieval.
Zhang et al.~\cite{Zhang_2017_ICCV} develop a range loss to effectively utilize the long tail training data.
Pose invariant representation is the key step for real world robust recognition system,
and has been the focus of many works. 
For example, Masi et al.~\cite{Masi_2016_CVPR} propose the face representation by fusing multiple pose-aware CNN models.
Peng et al.~\cite{Peng_2017_ICCV} untangle the identity and pose in representation by reconstruction in the feature space. 
Lu et al.~\cite{Lu_2017_WACV} propose the joint optimization framework for face and pose tasks.

\noindent \textbf{Face Frontalization and Canonicalization.}
Prior works in face frontalization and canonicalization optimize an image warping to fit a 3D face model~\cite{Zhu_2015_CVPR}, \cite{Hassner_2015_CVPR} based on localized 2D facial landmarks.
Recently, several attempts have been made to improve the generated face quality with neural networks.
Early works~\cite{Zhu_2013_ICCV},\cite{zhu2014multi} calibrate faces of various poses into canonical view, and disentangle the pose factor from identity with convolution neural networks.
Yim et al.~\cite{Yim_2015_CVPR} improve the identity preserving ability by introducing an auxiliary task to reconstruct the input data.
Cole et al.~\cite{Cole_2017_CVPR} decompose the generation module into geometry and texture parts, training with the differentiable warping. 

Recent works further improve the generation quality with Generative Adversarial Network (GAN)~\cite{goodfellow2014generative}.
Tran et al.~\cite{Tran_2017_CVPR} propose DR-GAN to simultaneously learn the frontal face generation and discriminative representation disentangled from pose variations.
Yin et al.~\cite{Yin_2017_ICCV} introduce a 3DMM reconstruction module in the proposed FF-GAN framework to provide better shape and appearance prior.
Huang et al.~\cite{Huang_2017_ICCV} incorporate both global structure and local details in their generator with landmark located patch networks. 
In our method, we do not require frontal and profile training pairs that are needed in the previous work,
and our rectification process is recognition oriented, which induces better recognition performance. 

\noindent \textbf{Spatial Transformer Network.}
The Spatial Transformer Network (STN)~\cite{jaderberg2015spatial} performs spatial transforms in the image or feature maps with a differential module, which can be integrated into the deep learning pipeline and optimized end-to-end .
The most relevant application of STN to our work is image alignment.
Kanazawa et al.~\cite{Kanazawa_2016_CVPR} match the fine-grained objects by establishing correspondences between two input images with non-rigid deformations.
Chen et al.~\cite{chen2016supervised} use STN to warp face proposals to canonical view with detected facial landmarks.
Zhong et al.~\cite{7948720} use STN for face alignment before recognition.
Lin et al.~\cite{Lin_2017_CVPR} provide a theoretical connection between STN and the Lucas-Kanade algorithm, and introduce the the inverse composition STN to reduce input variations.

The recent work Wu et al.~\cite{Wu_2017_ICCV} propose a recursive spatial transformer (ReST) for the alignment-free face recognition. 
They also integrate the recognition network in an end-to-end optimization manner.
There are two major differences between our approach and ReST. 
First, instead of manually dividing the facial region into several regions to allow non-rigid transformation modeling, we use a group of deformable plane cells to deal with complex warping effects. Second, we introduce a regularization prior of canonical view face to achieve better rectification effects. 

\section{Approach}
 \noindent {\bf Notation.}
 Let $I^X, I^Y$ denote the original image and rectified image. 
 We define the coordinate in the original image $I^X$ as the \emph{original coordinate}, and the one in the rectified image $I^Y$ as the \emph{rectified coordinate}. 
 Let $p=[p_x, p_y]^T$ and $q=[q_x, q_y]^T$ denote the points in the original coordinate and rectified coordinate. 
 We use $\hat{p}$ and $\hat{q}$ to denote the homogeneous coordinates as $\hat{p}=[p_x, p_y, 1]^T, \hat{q}=[q_x, q_y, 1]^T$. 
 Without loss of generality, we assume the coordinates of pixels are normalized to $[0,1)\times[0,1)$.
 
\subsection{Overview}\label{sec:overview}
The system contains two parts: the rectification module and the recognition module (Fig.~\ref{fig:overview}).
In the rectification process, the rectification network $f_\theta$ with parameter $\theta$ maps the original face image $I^X$ into the rectified one $I^Y$ by non-rigid image warping.
Then, the recognition network $g_\phi$ is trained with the metric learning based on the rectified image $I^Y$.
We further introduce a regularization to encourage the rectified face in canonical views, which is modeled as a prior under the distribution of natural faces with canonical views.

\subsection{Face Rectification Network} \label{sec:warp}
\begin{figure}[t]
\floatsetup{capbesideposition=right}
\fcapside[\FBwidth]{\caption{ {\bf Local Homography Transformation.}
  The rectification process approximates the 3D face as plane cells and canonicalizes it with local homographies. The rectified image is partitioned into $n^2$ cells, and the corresponding homographies are estimated by the rectification network. We put springs at the corners of the cells as soft constraints to avoid large discontinuities in the boundaries.
  }}{\includegraphics[width=1.0\linewidth]{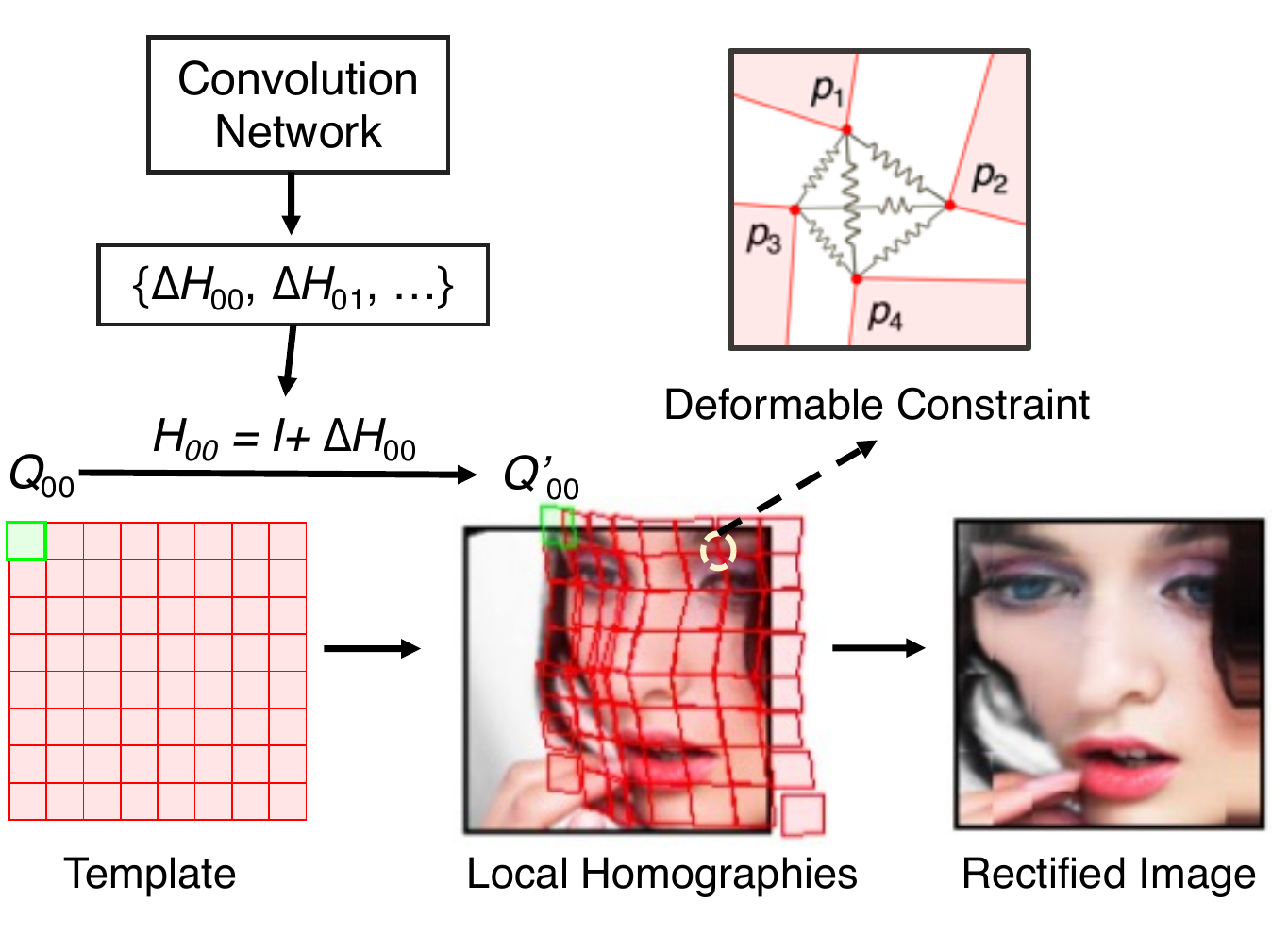}\label{fig:transformation}}
\end{figure}

In this section, we present the rectification process.
Different from recent face frontalization techniques~\cite{Tran_2017_CVPR,Yin_2017_ICCV,Huang_2017_ICCV} generating faces from abstract feature, 
 we define the rectification process as warping pixels from the original image to the canonical one, as illustrated in Fig.~\ref{fig:transformation}.
 
Formally, we define a template $Q$ by partitioning the rectified image into $n^2$ non-overlapped cells
 \begin{align}
 &Q=\big\{Q_{i,j}\big\}, 0 \le i,j < n, \notag \\
 &Q_{i,j} = \bigg[\frac{i}{n}, \frac{i+1}{n}\bigg)\times \bigg[\frac{j}{n}, \frac{j+1}{n}\bigg).
 \end{align}
 For each cell $Q_{i,j}$, we compute the corresponding deformed cell $Q'_{i,j}$ in the original image by
 estimating a local homography $H_{i,j}$.
 
 Specifically, we formulate the homography matrix as
 \begin{align}
 H_{i,j} = \begin{pmatrix}
 1+h_1 & h_2 & h_3 \\
 h_4 & 1+h_5 & h_6 \\
 h_7 & h_8 & 1
 \end{pmatrix} = I + \Delta H_{i,j}
 \end{align}
 The rectification network takes the original image $I^X$ as input and predicts $n^2$ residual matrices $\Delta H_{i,j}$.
 Then the rectified image $I^Y$ at cell $Q_{i,j}$ is obtained with homographies $H_{i,j}=I+\Delta H_{i,j}$ as 
 \begin{align}
 I^Y_q = I^X_p, q \in Q_{i,j}, p \in Q'_{i, j}, \quad \text{s.t.} \quad \lambda \hat{p} = H_{i,j}\hat{q}, \lambda \neq 0,
 \end{align}
 where $\hat{p},\hat{q}$ are the homogeneous coordinates of $p,q$.

Let $C$ denote the collection of corners of each cell $Q_{i,j}$ as $C = \{(\frac{i}{n}, \frac{j}{n}); 0\le i,j\le n\}$.
Since all the local homographies are estimated separately, a cell corner $c_i \in C$ in the rectified image is mapped to multiple points in the original image (see Fig.~\ref{fig:transformation}). 
In order to avoid large discontinuities between the boundaries of neighboring cells in $I_X$, we further introduce a soft constraint, called deformation constraint $\mathcal{L}_{\text{de}}$.
Specifically, let $M_i$ denotes the collection of $c_i$'s mapping coordinates in the original image. 
Then a soft constraint $\mathcal{E}_{c_i}$ is added to enforce the conformity between every pair of points in $M_i$ as
$
\mathcal{E}_{c_i} = \sum_{u,v \in M_i} ||u-v||_2
$.
We incorporate this soft constraint into the learning objective, 
and cast it as the the deformation loss of the rectification network:
\begin{align}
\mathcal{L}_{de} = \frac{1}{|C|} \sum_{c_i\in C} \mathcal{E}_{c_i}.
\end{align} 

\subsection{Regularization by Denoising Autoencoder} \label{sec:prior}
The regularization encourages that the rectification process generates face in canonical views.
We define it as an image prior that is directly based on the natural canonical view face distribution $P_Y$ as
\begin{equation}
\mathcal{L}_{reg} = -\log P_Y(I^Y).\label{eq:prior}
\end{equation}
In general, this optimization is non-trivial. We do not explicitly model the distribution, 
but consider the gradient of $\log P_Y$ and maximize it with stochastic gradient descent
\begin{align}
 \frac{\partial}{\partial \theta} \log P_Y(I^Y) &=
\frac{\partial}{\partial I^Y} \log P_Y(I^Y)  \frac{\partial I^Y}{\partial \theta} .
\end{align}
Using results from~\cite{sarela2005denoising},\cite{alain2014regularized}, 
which are also used in image generation~\cite{Nguyen_2017_CVPR} and restoration~\cite{sonderby2014apparent},\cite{bigdeli2017deep},\cite{Meinhardt_2017_ICCV}, 
we approximate the gradient of the prior as
\begin{align}
\frac{\partial}{\partial I^Y} \log P_Y(I^Y) &\approx \frac{h_{\omega^*}(I^Y)-I^Y}{\sigma^2}.\label{eq:approx}
\end{align}
Here $
h_{\omega^*} =  \argmin_{h_\omega} E_{y} || h_\omega(y+\sigma\epsilon)-y||^2_2 $, with 
$\epsilon \sim N(0, I)$ and $y \sim P_Y$, 
is the optimal denoising autoencoder trained on the true data distribution $P_Y$ (canonical view faces in our work)
with the infinitesimal noise level $\sigma$.
Using these results, we optimize the Eqn.~\ref{eq:prior} by first training a Denoising Autoencoder $h_\omega$ on the canonical view face dataset, and then estimating the approximated gradient in backpropagation via Eqn.~\ref{eq:approx}.

\subsection{Face Recognition Network} \label{sec:recog}
Given the rectified face $f_\theta(I^X)=I^Y$, we extract the face representation $g_\phi(I^Y)$ with deep convolutional recognition network $g_\phi$.
Following the previous works~\cite{Schroff_2015_CVPR}, 
we train the recognition network with triplet loss.
Let $D=\{I^X_o, I^X_+, I^X_-\}$ denote the three images forming a face triplet where $I^X_o$ and $I^X_+$ are from the same person, while $I^X_-$ is from a different person.
the recognition loss is
\begin{align}
\mathcal{L}_{recog} =  \max\bigg(0, d\big(I^X_o, I^X_+\big) -d\big(I^X_o, I^X_-\big)+\alpha\bigg).
\end{align}
where $d(x,y) = ||g_\phi(f_\theta(x))-g_\phi(f_\theta(y))||_2$ is the Euclidean distance between the feature representations $x$ and $y$. 
The hyper-parameters $\alpha$ control the margin between intra-person distance and inter-person distance in the triplet loss.

In summary, we jointly optimize the rectification network and recognition network by minimizing an objective, consisting of a deformable term, a recognition term, and a regularization term
\begin{equation}
\argmin_{\theta,\phi} \,\,\,\,  \mathcal{L}_{\text{recog}} + \lambda_{\text{reg}} \mathcal{L}_{\text{reg}} 
+ \lambda_{\text{de}} \mathcal{L}_{\text{de}}.\label{eq:obj}
\end{equation}

 \begin{table}[t]
\centering
\footnotesize
\begin{tabular}{p{2cm} | l | l}
\hline
Network & {\bf Rectification Network} & \bf Denoising Autoencoder  \\\hline
Input & \multicolumn{2}{c}{$128\times128\times3$} \\\hline
Stage-1 & \tabincell{l}{Conv[8, 3, 2, 1] \\ MaxPool[2,2,1] \\ Conv[32, 3, 2, 1]} & 
\multirow{1}{27mm}{Conv[8, 3, 2, 1] \\
Conv[12, 3, 2, 1]\\
Conv[16, 3, 2, 1]\\
Conv[24, 3, 2, 1]\\
FullyConnected[1536]\\
DeConv[24, 3, 2, 1]\\
DeConv[16, 3, 2, 1]\\
DeConv[12, 3, 2, 1]\\
DeConv[8, 3, 2, 1]\\
Conv[3, 3, 1, 1]
}\\\cline{1-2}
Stage-2 & \tabincell{l}{InceptionModule[16] \\ MaxPool[2, 2]} & \\\cline{1-2}
Stage-3 & \tabincell{l}{InceptionModule[32]*2 \\ MaxPool[2, 2]} & \\\cline{1-2}
Stage-4 & \tabincell{l}{InceptionModule[64]*2 \\ MaxPool[2, 2]} & \\\cline{1-2}
Stage-5 & \tabincell{l}{FullyConnected[128] \\ FullyConnected[N]} & \\\hline
\end{tabular}
\caption{{\bf Network Details.}
Conv[$ch, w, s, p$] denotes a convolution layer with kernel size $ch\times w \times w$, stride $s$ and padding $p$. 
The deconvolution layer DeConv[$ch,w,s,p$] is implemented as the gradient of convolution with respect to data,
and the meaning of parameters is still in a convolution sense.
MaxPool[$w,s$] is a max-pooling layer with $w\times w$ window and stride $s$.
FullyConnected[$n$] is a fully-connected layer with $n$ output neurons, 
and $N$ denotes the number of corresponding transformation parameters.
InceptionModule[$ch$] denotes a modified Inception module with the same number of feature maps $ch$ in each branch.
}\label{tab:network}
\end{table}

 \begin{figure*}[t]
\includegraphics[width=0.9\linewidth]{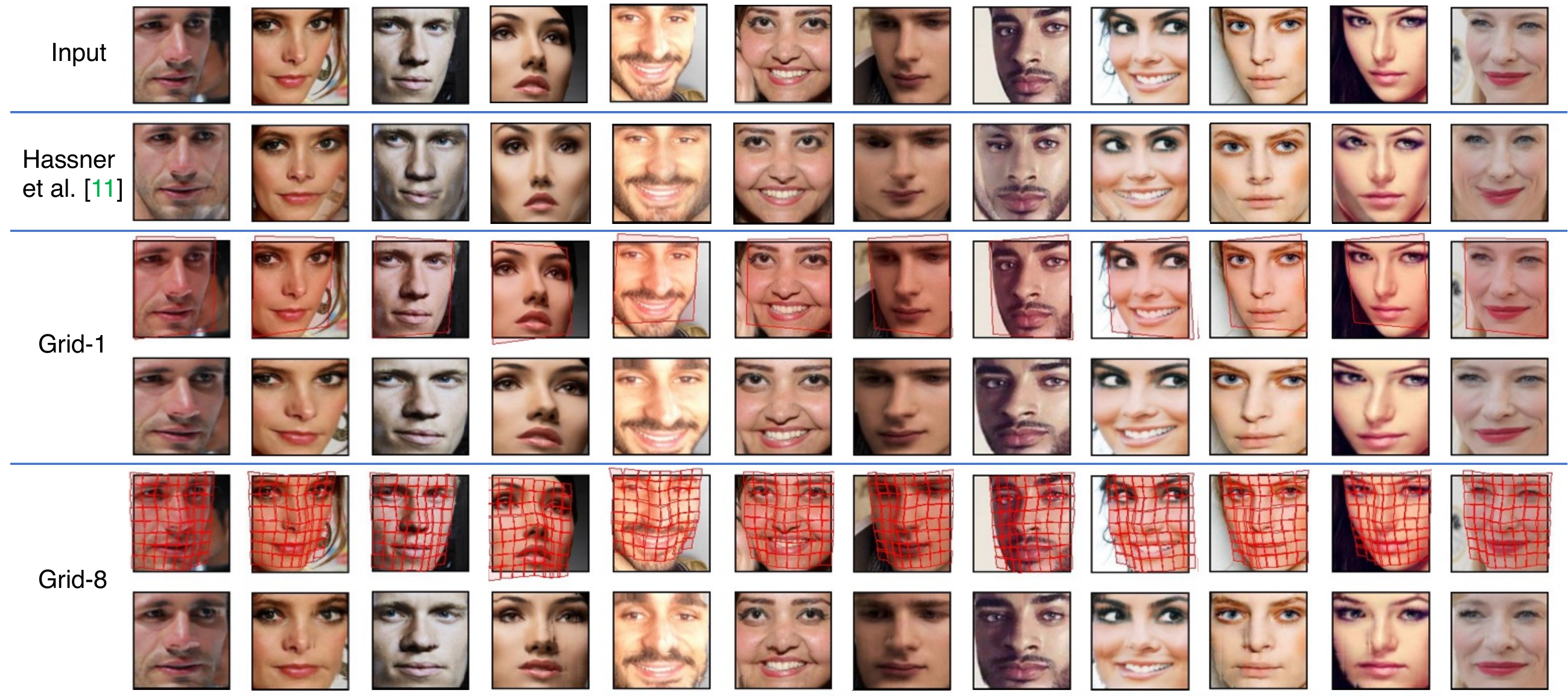}
   \caption{
   \textbf{Qualitative Analysis of SNFace Testset.} 
   We sample the data from the SNFace test set with pose, expression, and illumination variations, and visualize the rectified results under different rectification methods.
   }
   \label{fig:comparison}
\end{figure*} 

\section{Experiments}\label{sec:exp}

\subsection{Experimental Details}\label{details}
\noindent\textbf{Dataset.}
Our models are learned from a large collection of photos in social networks, referred to as the Social Network Face Dataset (SNFace).
The SNFace dataset contains about 10M images and 200K individuals.
We randomly choose 2K individuals as the validation set, 2K as the test set, and use the remaining ones as the training set.
The 5-point facial landmarks are detected and the face is aligned with similarity transformation.
  
\noindent\textbf{Network Architecture.}
In all the experiments in this paper, we use the GoogLeNet~\cite{Szegedy_2015_CVPR}, \cite{Schroff_2015_CVPR} for our recognition network.
The rectification network is based on a modified Inception module, which contains fewer parameters and a simpler structure.
The rectification network takes very limited additional parameters and time computation compared with the recognition network.
The Denoising Autoencoder is designed with a convolutional autoencoder structure.
The network details are described in Tab.~\ref{tab:network}.

\noindent\textbf{Implementation Details.}
The dimension of the original and rectified face of the rectification network are $128\times128$, 
and the pixel level activations are normalized by dividing 255.
The Denoising Autoencoder is trained on a subset of the SNFace dataset, which contains 100K faces in canonical views.
An end-to-end optimization is conducted after the Denoising Autoencoder is ready.
In the training phase, each mini-batch contains 1024 image triplets.
We set an equal learning rate for all trainable layers to 0.1, 
which is shrunk by a factor of 10 once the validation error stops decreasing.
The hyper parameters are determined by the validation set as $\lambda_{\text{reg}}=10.0,\alpha=0.3$, and $\lambda_{\text{de}}=1.0$.
In all the experiments, we use the same metric learning method with triplet loss. 
No other data processing and training protocol are used.
In the testing phase, we use the Euclidean distance as the verification metric between two face representations.

\subsection{What is Learned in Face Rectification?}\label{sec:comparison}
In this section, we study what is learned in the rectification network.
All approaches are evaluated on the SNFace test dataset.
We evaluate our model with $n=8$ (i.e., 64 cells in local homography transformations), referred to as \emph{Grid-8}.
We compare with several alternative approaches: 
the \emph{baseline} model does not have face rectification; 
the global model \emph{Grid-1} performs the face rectification with global homography transformation;
no face prior regularization model \emph{Grid-8\textbackslash reg} does not have the regularization during training.

Moreover, in order to compare with the 3D face frontalization technique used in face recognition (e.g., 3D alignment used in DeepFace~\cite{Taigman_2014_CVPR}),
we process the full SNFace dataset to synthesize frontal views by using a recent face fronalization method created by Hassner et al.~\cite{Hassner_2015_CVPR},
and compare with the model trained on this synthesized data (called \emph{baseline-3D}) to verify the effectiveness of our rectification process and joint optimization.
\begin{table}[t]
\begin{center}
\begin{tabular}{p{2cm} p{2cm} p{2cm} p{2cm}}
\hline
\multicolumn{4}{c}{\bf Evaluation on SNFace Testset}\\
Method$\,\downarrow$& FAR=$10^{-2}$& FAR=$10^{-3}$ & FAR=$10^{-4}$ \\\hline
baseline & 92.94 & 81.76 & 63.41\\\hline
baseline-3D & 94.02 & 80.36 & 58.20 \\\hline
Grid-1 & 93.49 & 83.94 & 66.15\\
Grid-2 & 94.02 & 85.24 & 68.70\\
Grid-4 & 94.38 & 86.23 & 71.09\\\hline
Grid-8\textbackslash reg & 94.10 & 85.44 & 69.05\\
Grid-8 & \bf 94.92 & \bf 87.81 & \bf 72.71\\\hline
\end{tabular}
\end{center}
\caption{{\bf Quantitative Results on the SNFace Testset.}
We compare our method \emph{Grid-8} against several other approaches and report verification accuracy on the SNFace test set.
}\label{tab:comparison}
\end{table}

\noindent{\bf Qualitative Analysis.}
Fig.~\ref{fig:comparison} depicts the visualization results of the original images and the corresponding rectified images.
Obviously, the global homography transformation \emph{Grid-1} can capture coarse geometric information, such as 2D rotation and translation, which is also reported in previous works~\cite{7948720},\cite{Wu_2017_ICCV}. 
However, due to its limited capacity, \emph{Grid-1} is unable to satisfactorily rectify out-of-plane rotation and local geometric details, 
and generate with notable distortion (e.g., the big nose in the faces with large pose).
Hassner et al.~\cite{Hassner_2015_CVPR} improve further, generating good frontal view faces, but the ghosting effect (most faces under large pose in Fig.~\ref{fig:comparison}) and the change of facial shape (e.g., nose in the fourth individual in Fig.~\ref{fig:comparison}) may introduce further noise to the recognition system.
On the other hand, \emph{Grid-8} can capture rich facial structure details.
Local geometric warping is detected and approximated by local homographies.
Compared with the original image and results from other approaches, 
the proposed method \emph{Grid-8} greatly reduces geometric variations and induces better canonical view results.

\noindent{\bf Quantitative Analysis.}
We report quantitative results under verification protocol in Tab.~\ref{tab:comparison}.
\emph{Grid-8} achieves the best performance which outperforms the baseline by a large margin up from $63.4\%$ to $72.7\%$ with False Alarm Rate (FAR) at $10^{-4}$.
The global transformation \emph{Grid-1} consistently improves the recognition performance compared with the \emph{baseline}.
But as we have seen in the visualization results, global transformation is limited to its transformation capacity and thus introduces large distortion for recognition. 

The recognition model trained on the synthesized frontal view data \emph{baseline-3D} obtains high performance with FAR at $10^{-2}$, better than the \emph{baseline} and \emph{Grid-1} trained on the original data.
But the performance drops dramatically, and finally gets $5.2\%$ worse than the \emph{baseline} with FAR at $10^{-4}$.
On the other hand, our method \emph{Grid-8} consistently outperforms the \emph{baseline-3D} and obtains $14.5\%$ improvement with FAR at $10^{-4}$.

\begin{figure}[t]
\floatsetup{capbesideposition=right}
\fcapside[\FBwidth]{\caption{
\textbf{Synthetic 2D Transformations.} Visualization of the image and perturbed samples in the synthetic 2D transformation experiment. 
   (a). Original image, where different color boxes corresponding to different noise levels (red for $\sigma=0.05d$, green for $\sigma=0.1d$, and blue for $\sigma=0.15d$).
   (b). Cropped faces with noisy landmarks.
   (c). Rectified faces by our method \emph{Grid-8}. Most of the scale, rotation, and translation variations are reduced.
   }
}{\includegraphics[width=1.0\linewidth]{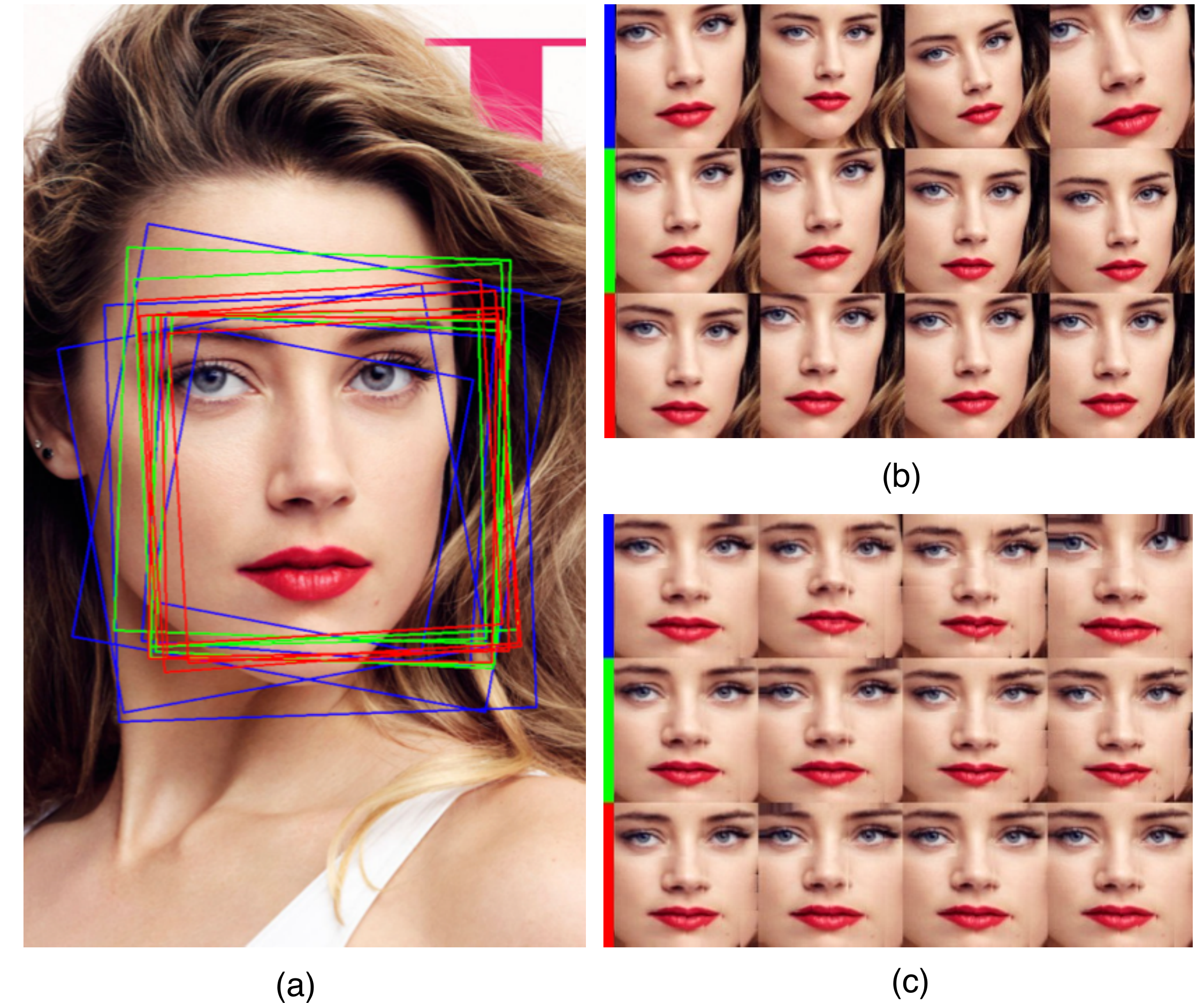}\label{fig:align}}
\killfloatstyle
\begin{floatrow}
\floatsetup{capbesideposition=inside}
\ttabbox[\Xhsize]{\caption{{\bf Quantitative Results under Synthetic 2D Transformations.}
Verification accuracy of our model \emph{Grid-8} and \emph{baseline} at FAR=$10^{-2}$ 
under 2D transformations with different noise levels.
}\label{tab:align}}{
\begin{tabular}{p{2cm} p{2cm} p{2cm} p{2cm} p{2cm}}
\hline
Method$\,\downarrow$& $\sigma=0.00d$ & $\sigma=0.05d$ & $\sigma=0.10d$ & $\sigma=0.15d$ \\\hline
baseline & 92.94 & 91.66 & 86.58 & 74.95\\\hline
Grid-8 & 94.92 & 93.51 & 90.35 & 85.00\\\hline\\
\end{tabular}
}
\end{floatrow}

\includegraphics[width=1.0\linewidth]{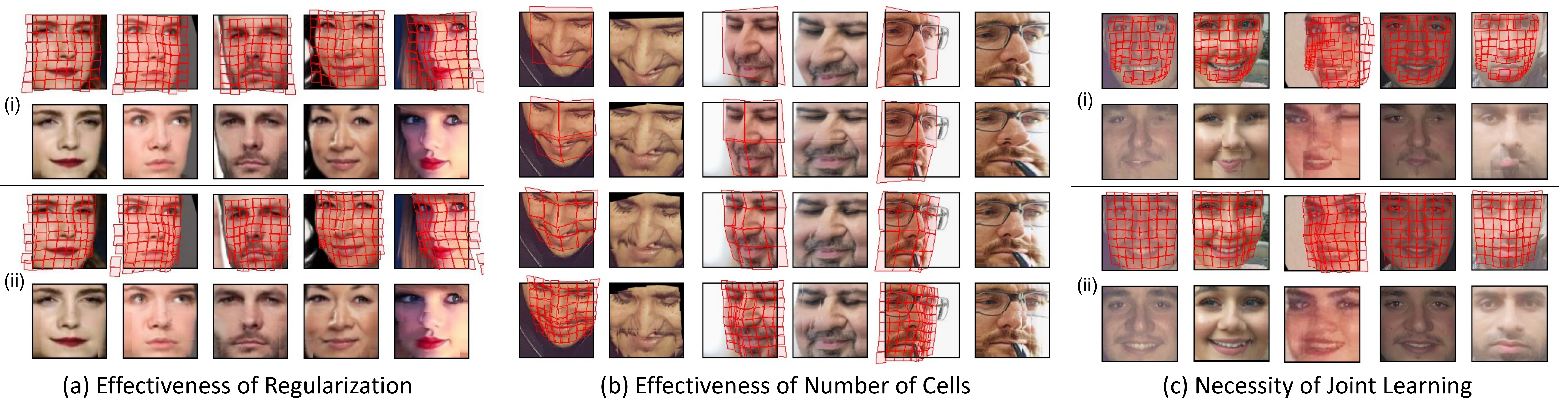}
\caption{
\textbf{Ablation Studies.} 
(a) i. Rectification without regularization. ii. Rectification with regularization. 
(b) Rectification with different number of cells.
(c) i. Rectification without recognition supervision. ii. Joint learning of rectification and recognition.
   }\label{fig:ab}
\end{figure}

\noindent{\bf Evaluation on Synthetic 2D Transformations.}
We investigate the effectiveness of face rectification for reducing 2D in-plane transformations, which is typically introduced by facial landmarks.
The perturbed data is generated by performing face alignment with noisy landmarks, which are synthesized by adding \emph{i.i.d.} Gaussian noise of variance $\sigma$.
The Gaussian noise mimics the inaccurate facial landmarks in the real system, and introduces the scale, rotation, and translation variations in the face alignment.
We normalize the face size by interocular distance $d$, and generate perturbed data with different noise levels $\sigma=0.05d, 0.1d, 0.15d$.

Fig.~\ref{fig:align} presents the visualization of synthetic data with small (red boxes with $\sigma=0.05d$), 
middle (green boxes with $\sigma=0.1d$) and large (blue boxes with $\sigma=0.15d$) noise levels.
As shown in Fig.~\ref{fig:align} (c), the rectification network can generate canonical view faces, which greatly reduce the in-plane variance.
Tab.~\ref{tab:align} reports the qualitative comparison between the \emph{baseline} and \emph{Grid-8}.
We can see that the \emph{baseline} suffers from the large in-plane variations and the accuracy drops rapidly.
Meanwhile, the rectification network \emph{Grid-8} yields much better performance even under large geometric variations.

\noindent{\bf Effectiveness of Regularization.}
We further explore the effectiveness of regularization.
Visualization results of rectified faces are shown in Fig.~\ref{fig:ab}(a).
The first two rows present the rectification trained without regularization, and the last two rows show the results with regularization.
We can observe that the regularization helps the rectification process generate more canonical view faces, and reduces the cropping variations in the rectified results.
Quantitative results are reported in Tab.~\ref{tab:comparison}.
The regularization achieves $2.4\%$ improvement with FAR at $10^{-3}$, 
and $3.7\%$ improvement with FAR at $10^{-4}$.

 \begin{figure}[t]
 \centering
\includegraphics[width=0.8\linewidth]{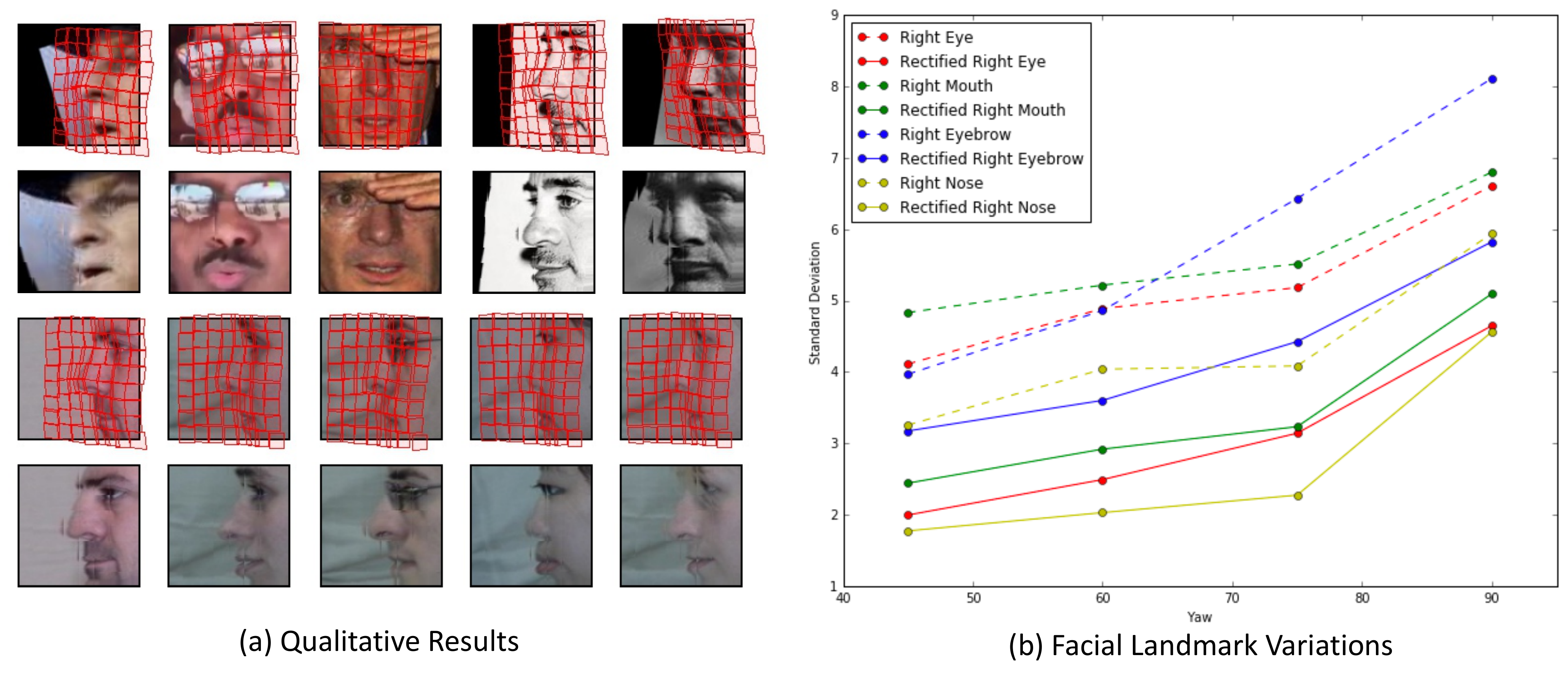}
   \caption{
   \textbf{Evaluation on Challenge Situations.}
   (a). Qualitative results under large pose and occlusion.
   (b). Comparisons of standard deviation of facial landmarks under different pose variations.
   }
   \label{fig:hard}
\end{figure} 

\noindent{\bf Number of Partition Cells.}
We investigate the influence of the number of partition cells in the rectification network.
Visualized results of $n=1, 2, 4, 8$ are presented in Fig.~\ref{fig:ab}(b),
and the quantitative results in SNFace test set are shown in Tab.~\ref{tab:comparison}.
As the number of cells increases, the image distortion introduced in the rectified face decreases,
 and verification performance increases, benefiting from the local homography transformations. 

\noindent{\bf Necessity of Joint Learning.}
To evaluate the contribution of joint learning the face rectification and recognition,
we introduce an ablation experiment learning each part sequentially.
This model first learns the face rectification without the recognition supervision, and then trains the recognition module with the fixed rectification module. 
Fig.~\ref{fig:ab}(c) provides the qualitative results. 
The consequences of the lack of recognition supervision is obvious and irreversible.
The noisy gradient provided by the Denoising Autoencoder introduces much artifacts and the misalignment objective further drops many face details (e.g. close the mouth in the first and second individuals).
On the other hand, joint learning of rectification and recognition can greatly reduce artifacts in the rectification results and keep the most of facial details.  
The recognition accuracy of this sequential learning model is $91.1\%, 72.3\%, 41.6\%$ with FAR at $10^{-2}, 10^{-3}, 10^{-4}$,
which is far below the joint learning model and even the original baseline.

\noindent{\bf Evaluation on Challenge Situations.}
Fig.~\ref{fig:hard}(a) presents the rectification results under challenging occlusion situations like large pose and sunglasses. 
The effects of rectification process is not hallucinating the missing parts. 
It reduces the geometric variations and does alignment for the visible parts. 
We further evaluate the variations of facial landmarks on the Multi-PIE dataset~\cite{gross2010multi}.
Four facial landmarks in the right side of face are considered and the corresponding standard deviations are calculated.
Fig.~\ref{fig:hard}(b) demonstrates the landmark variations in the original and rectified faces under different face pose.
Obviously, the variations of each landmark across different poses are much smaller than ones in the original face,
which suggests that our rectification process is robust to pose variation and reduce facial geometric variations significantly.

\subsection{Evaluation on Public Benchmarks}~\label{sec:public}
To verify the cross-data generalization of learned models, we report our performance on four challenge public benchmarks, which cover large pose, expression, and illumination variations.
We further report our models trained under the public dataset MS-Celeb-1M~\cite{guo2016ms}, referred to as \emph{baseline-Pub} and \emph{Grid-8-Pub}.

\noindent {\bf LFW and YTF}. 
In the LFW dataset~\cite{LFWTech}, we follow the standard evaluation protocol of unrestricted with labeled outside data, 
and report the mean accuracy (mAC) of 10-folders verification set. 
We further follow the identification protocol proposed by Best-Rowden et al.~\cite{best2014unconstrained}, and report the closed-set recognition performance measured by rank-1 rate and the open-set performance measured by Detection and Identification Rate (DIR) with FAR at $1\%$.
In the YTF dataset~\cite{wolf2011face}, we follow the standard protocol and report the mAC of 10 folds video verification set.
We perform the video-to-video verification by averaging the similarity scores between every pairs of images.

 \begin{figure}[t]
 \centering
\includegraphics[width=0.8\linewidth]{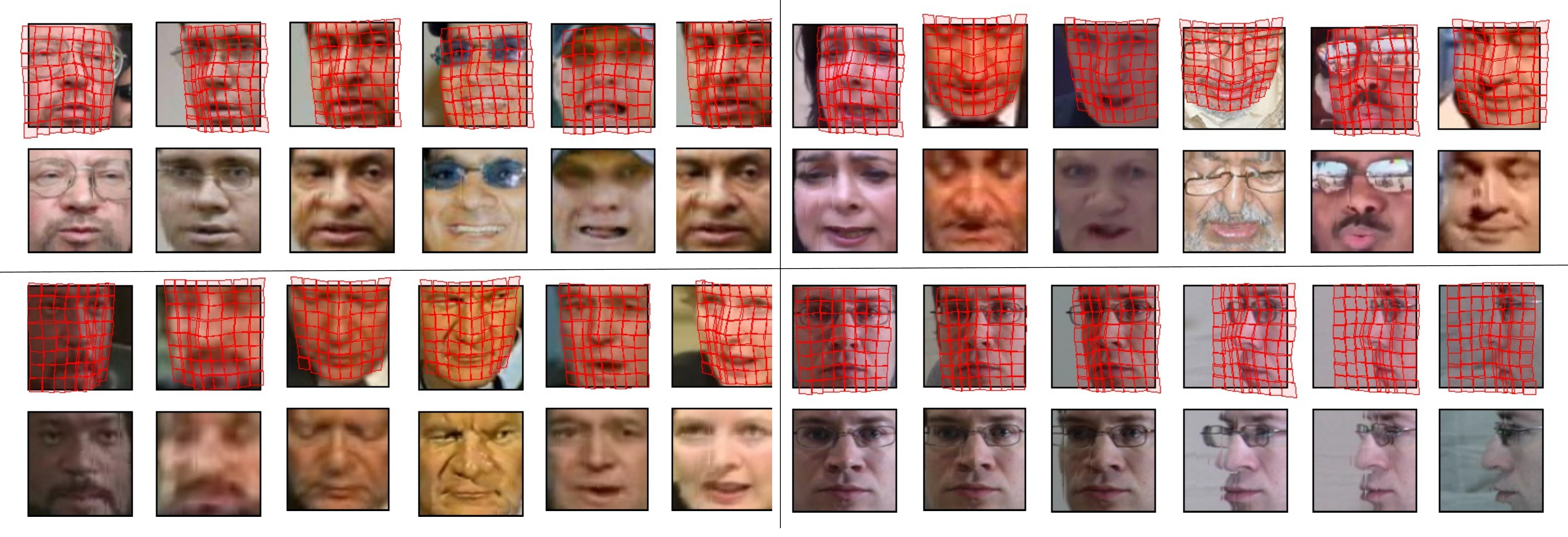}
   \caption{
   \textbf{Qualitative Analysis on Public Benchmarks.} 
   Left Top: LFW; Left Bottom: YTF;
   Right Top: IJB-A; Right Bottom: Multi-PIE.
   }
   \label{fig:bm-res}
\end{figure}

\noindent {\bf Results on LFW and YTF.} 
Tab.~\ref{tab:eva-lfw-ytf} shows our results. 
In the LFW verification benchmark, 
our method consistently improves the performance up from $99.05\%$ to $99.68\%$ with the MS-Celeb training set and from $99.15\%$ to $99.70\%$ with the SNFace training set. 
Our results are comparable with FaceNet~\cite{Schroff_2015_CVPR} but with the considerably smaller training data (10M training faces VS 200M faces). 
Under the LFW identification protocol, 
our method boosts the baseline with significant improvements (up from $91.7\%$ to $96.7\%$ in the close-set protocol and from $80.3\%$ to $94.1\%$ in the open-set protocol), 
and achieves the state-of-the-art.
In the YTF benchmark, our method \emph{Grid-8} ($95.6\%$) and \emph{Grid-8-Pub} ($95.2\%$) also provide consistent improvements over the baseline methods \emph{baseline} ($94.0\%$) and \emph{baseline-Pub} ($93.4\%$).
Fig.~\ref{fig:bm-res} provides the rectification results in LFW and YTF. 

\noindent {\bf Multi-PIE}. 
The Multi-PIE dataset~\cite{gross2010multi} contains 754,200 images from 337 subjects, covering large variations in pose and illumination. 
We follow the protocol from~\cite{Yim_2015_CVPR}, where the last 137 subjects with 13 poses, 20 illuminations and neutral expression are selected for testing. 
For each subject, we randomly choose one image with frontal pose and neutral illumination as the gallery, and leave all the rest as probe images.

\noindent {\bf Results on Multi-PIE.} 
Tab.~\ref{tab:eva-multipie} shows our results.
Our method outperforms the baseline methods by a large margin, from $44.3\%$ (\emph{baseline-Pub}) to $62.0\%$  (\emph{Grid-8-Pub}) and $65.5\%$ (\emph{baseline}) to $75.4\%$ (\emph{Grid-8}) in the identification rate with face yaw at $90^\circ$ .
We achieve the best performance and outperforms the recent GAN-based face frontalization methods~\cite{Tran_2017_CVPR,Huang_2017_ICCV,Yin_2017_ICCV}.
Moreover, we do not observe the performance degeneration in frontal face, which indicates that our method introduces few artifacts in frontal faces and gains consistent improvement over pose variations.
Fig.~\ref{fig:bm-res} shows the qualitative results with different pose variations. 
The rectification process is robust to the change of pose, and reduces the geometric variations for the visible parts.

\noindent {\bf IJB-A.} 
The IJB-A dataset is a challenge benchmark due to its unconstrained setting.
It defines the set-to-set recognition as face template matching.
In our evaluation, we do not introduce complicate strategies, and perform the set-to-set recognition via a media pooling followed from the previous method~\cite{sankaranarayanan2016triplet}.
Specifically, the template feature is extracted by first averaging all image feature with their media ID, and then averaging between medias.

\noindent {\bf Results on IJB-A.} 
Tab.~\ref{tab:eva-ijba} and Fig.~\ref{fig:bm-res} report our results on IJB-A.
It is worth pointing out that we employ strong baselines, which achieve $68.5\%$ (\emph{Baseline-Pub}) and $71.3\%$ (\emph{Baseline}) verification accuracy with FAR at 0.001, and $89.6\%$ (\emph{Baseline-Pub}) and $90.4\%$ (\emph{Baseline}) rank-1 identification accuracy.
By adding our rectification process, our rectification methods outperform the these strong baselines by a large margin.
We achieve $13.8\%$ (\emph{Grid-8-Pub}) and $12.6\%$ (\emph{Grid-8}) improvement on the verification task (with FAR at 0.001),
and reduce $25\%$ (\emph{Grid-8-Pub}) and $26\%$ (\emph{Grid-8}) error rate on the rank-1 identification task.
It is noteworthy that multiple-frame aggregation methods~\cite{Yang_2017_CVPR}, \cite{Liu_2017_CVPR} in the set-to-set recognition scenarios (e.g., IJB-A and YTF) can achieve better performance.
These techniques could also apply to our method and is left to the future work. 

\begin{table}[!t]
\centering
\ttabbox{\caption{Evaluation on LFW and YTF}\label{tab:eva-lfw-ytf}}{
\begin{tabular}{lcccc}
\hline
\multirow{1}{23mm}{Method$\,\downarrow$} & \multirow{1}{20mm}{LFW mAC} & \multirow{1}{23mm}{LFW Rank-1} & \multirow{1}{25mm}{LFW DIR@$1\%$} & \multirow{1}{20mm}{YTF mAC}\\\hline
DeepFace~\cite{Taigman_2014_CVPR} & 97.35 & 64.9 & 44.5 & 91.4\\
VGGFace~\cite{Parkhi15} & 99.13 & - & - &  97.4 \\
FaceNet~\cite{Schroff_2015_CVPR} &  99.64 & - & - & 95.1\\
DeepID2+~\cite{Sun_2015_CVPR} & 99.47 & 95.0 & 80.7 & 93.2\\
WST Fusion~\cite{Taigman_2015_CVPR} & 98.37 & 82.5 & 61.9 & - \\
SphereFace~\cite{Liu_2017_CVPR} & 99.42 & - & - & 95.0 \\
RangeLoss~\cite{Zhang_2017_ICCV} & 99.52 & - & - & 93.7 \\
HiReST-9+~\cite{Wu_2017_ICCV} & 99.03 & 93.4 & 80.9 & 95.4\\\hline
Baseline-Pub & 99.05 & 88.9 & 78.8 & 93.4 \\
Grid-8-Pub & 99.68 & 96.4 & 93.1 & 95.2 \\
Baseline & 99.15 & 91.7 & 80.3 & 94.0\\
Grid-8 & 99.70 &  96.7 &  94.1 & 95.6 \\\hline
\end{tabular}
}
\ttabbox{\caption{Evaluation on Multi-PIE}\label{tab:eva-multipie}}{
\begin{tabular}{p{2.55cm}p{1.2cm}p{1.2cm}p{1.2cm}p{1.2cm}p{1.2cm}p{1.2cm}p{1.2cm}}
\hline
Method$\,\downarrow$& $0^\circ$ & $15^\circ$ & $30^\circ$ & $45^\circ$ & $60^\circ$ & $75^\circ$ & $90^\circ$ \\\hline
Yim et al.~\cite{Yim_2015_CVPR} &99.5 & 95.0  & 88.5 & 79.9 & 61.9& - & -\\
DRGAN~\cite{Tran_2017_CVPR} & 97.0 & 94.0 & 90.1 & 86.2 & 83.2 & - & - \\
TPGAN~\cite{Huang_2017_ICCV} & - & 98.7 & 98.1 & 95.4 & 87.7 & 77.4 & 64.6 \\
FF-GAN~\cite{Yin_2017_ICCV} & 95.7 & 94.6 & 92.5 & 89.7 & 85.2 & 77.2 & 61.2  \\
\hline
Baseline-Pub & 100.0 & 100.0 & 100.0 & 98.9 & 92.9 & 78.4 & 44.3 \\
Grid-8-Pub & 100.0 & 100.0 & 100.0 & 99.3 & 96.1 & 86.7 & 62.0 \\
Baseline & 100.0 & 100.0 & 100.0 & 100.0 & 98.7 & 92.6 & 65.5\\
Grid-8 & 100.0 & 100.0 & 100.0 & 100.0 & 99.2 & 94.7 & 75.4\\\hline
\end{tabular}
}
\ttabbox{\caption{Evaluation on IJB-A}\label{tab:eva-ijba}}{
\begin{tabular}{p{2.5cm}p{2.2cm}p{2.2cm}p{2.2cm}p{2.2cm}}
\hline
Method$\,\downarrow$ & \multicolumn{2}{c}{Verification} & \multicolumn{2}{c}{Identification} \\
Metric$\,\rightarrow$ & @FAR=0.01& @FAR=0.001 & @Rank-1 & @Rank-5\\\hline
PAM~\cite{Masi_2016_CVPR} & $73.3 \pm 1.8$ & $55.2 \pm 3.2$ & $77.1 \pm 1.6$ & $88.7 \pm 0.9$\\
Masi et al.\cite{Masi2016} & $88.6 \pm 1.7$ & $72.5 \pm 4.4$ & $90.6\pm 1.3$ & $96.2 \pm 0.7 $ \\
TripEmbd.~\cite{sankaranarayanan2016triplet} & $90.0\pm 1.0 $ & $81.3\pm2.0$ & $93.2\pm1.0 $ & - \\
TempAdpt.~\cite{crosswhite2017template} & $93.9\pm1.3$ & $83.6\pm2.7$ & $92.8\pm1.0$ & $97.7\pm0.4$ \\
DRGAN~\cite{Tran_2017_CVPR} & $77.4\pm2.7$ & $53.9\pm4.3$ & $85.5\pm1.5$ & $94.7\pm1.1$ \\
FFGAN~\cite{Yin_2017_ICCV} & $85.2\pm1.0$ & $66.3\pm3.3$ & $90.2\pm0.6$ & $95.4\pm0.5$ \\\hline
Baseline-Pub & $86.6\pm1.8$ & $68.5\pm3.9$ & $89.6\pm1.3$ & $95.2\pm0.5$ \\
Grid-8-Pub & $91.5 \pm 0.8$& $82.3\pm 1.9$ & $92.2\pm 1.0$&$96.0\pm 0.5$ \\
Baseline & $88.7\pm1.9$ & $71.3\pm3.9$ & $90.4\pm1.1$ & $95.4\pm0.7$ \\
Grid-8 & $92.1\pm0.8$ & $83.9\pm1.4$ & $92.9\pm1.0$ & $96.2\pm0.5$ \\\hline
\end{tabular}
}
\label{tab:eva}
\end{table}

\section{Conclusion}
In this paper, we develop a method called \emph{GridFace} to reduce facial geometric variations.
We propose a novel non-rigid face rectification method by local homography transformations, 
and regularize it by imposing natural frontal face distribution with a Denoising Autoencoder.
Empirical results show our method greatly reduces geometric variations and improves the recognition performance.
\bibliographystyle{splncs04}
\bibliography{egbib}
\end{document}